%% file: main.tex
\documentclass[10pt,twocolumn,letterpaper]{article}

\usepackage{cvpr}              

\usepackage{graphicx}
\usepackage{amsmath}
\usepackage{amssymb}
\usepackage{booktabs}
\usepackage{multirow}
\usepackage{color}
\usepackage[table,xcdraw]{xcolor}
\usepackage{pifont}
\usepackage[T1]{fontenc}
\usepackage[pagebackref,breaklinks,colorlinks]{hyperref}

\usepackage[capitalize]{cleveref}
\crefname{section}{Sec.}{Secs.}
\Crefname{section}{Section}{Sections}
\Crefname{table}{Table}{Tables}
\crefname{table}{Tab.}{Tabs.}

\def\cvprPaperID{9082} 
\def\confName{CVPR}
\def\confYear{2023}

\begin{document}

\title{SE-ORNet: Self-Ensembling Orientation-aware Network  
for Unsupervised Point Cloud Shape Correspondence}

\author{Jiacheng Deng$^{1,}$\footnotemark[1]~, Chuxin Wang$^{1,}$\footnotemark[1]~, Jiahao Lu$^1$, Jianfeng He$^1$, Tianzhu Zhang$^{1,3,}$\footnotemark[2]~, Jiyang Yu$^2$, Zhe Zhang$^3$ \\
\small{$^1$University of Science and Technology of China, $^2$China Academy of Space Technology, $^3$Deep Space Exploration Lab}\\
{\tt\small \{dengjc, wcx0602, lujiahao, hejf\}@mail.ustc.edu.cn, {tzzhang@ustc.edu.cn}}\\
{\tt\small{yujiyang@spacechina.com}, {cnclepzz@126.com}}
}
\maketitle

\renewcommand{\thefootnote}{\fnsymbol{footnote}}
\footnotetext[1]{Equal Contribution}
\footnotetext[2]{Corresponding Author}
\begin{abstract}
   \input{abst.tex}
\end{abstract}

\vspace{-0.5em}
\section{Introduction}
\input{intro}

\section{Related Work}
\input{rw}

\section{Method}

\input{method}

\section{Experiments}
\input{expr}

\section{Conclusion}
\input{clu}

\section{Acknowledgements}
This work was partially supported by the National Nature Science Foundation of China (Grant 62022078, Grant 12150007) and National Defense Basic Scientific Research program of China (Grant JCKY2020903B002).

{\small
\bibliographystyle{ieee_fullname}
\bibliography{egbib}
}

\end{document}

%% file: abst.tex
Unsupervised point cloud shape correspondence aims to obtain dense point-to-point correspondences between point clouds without manually annotated pairs.
However, humans and some animals have bilateral symmetry and various orientations, which lead to severe mispredictions of symmetrical parts.
%
%
Besides, point cloud noise disrupts consistent representations for point cloud and thus degrades the shape correspondence accuracy.
To address the above issues, we propose a Self-Ensembling ORientation-aware Network termed SE-ORNet.
The key of our approach is to exploit an orientation estimation module with a domain adaptive discriminator to align the orientations of point cloud pairs, which significantly alleviates the mispredictions of symmetrical parts.
Additionally, we design a self-ensembling framework for unsupervised point cloud shape correspondence.
In this framework, the disturbances of point cloud noise are overcome by perturbing the inputs of the student and teacher networks with different data augmentations and constraining the consistency of predictions.
Extensive experiments on both human and animal datasets show that our SE-ORNet can surpass state-of-the-art unsupervised point cloud shape correspondence methods.
Code and demos are available at \url{https://github.com/OpenSpaceAI/SE-ORNet.git}.

%% file: intro.tex
%
With the cost of LiDAR and depth cameras falling, it is more accessible to obtain 3D point cloud data.
For real-world applications, such as articulated motion transfer~\cite{ding2015articulated, sun2022human} and non-rigid human body alignment~\cite{brown2007global}, the correspondence between two point clouds is indispensable.
However, we are hard to directly obtain the correspondence between two raw point clouds due to various object orientations and ununified coordinate systems. 

\begin{figure}[t]
    \centering
    \includegraphics[width=0.75\linewidth]{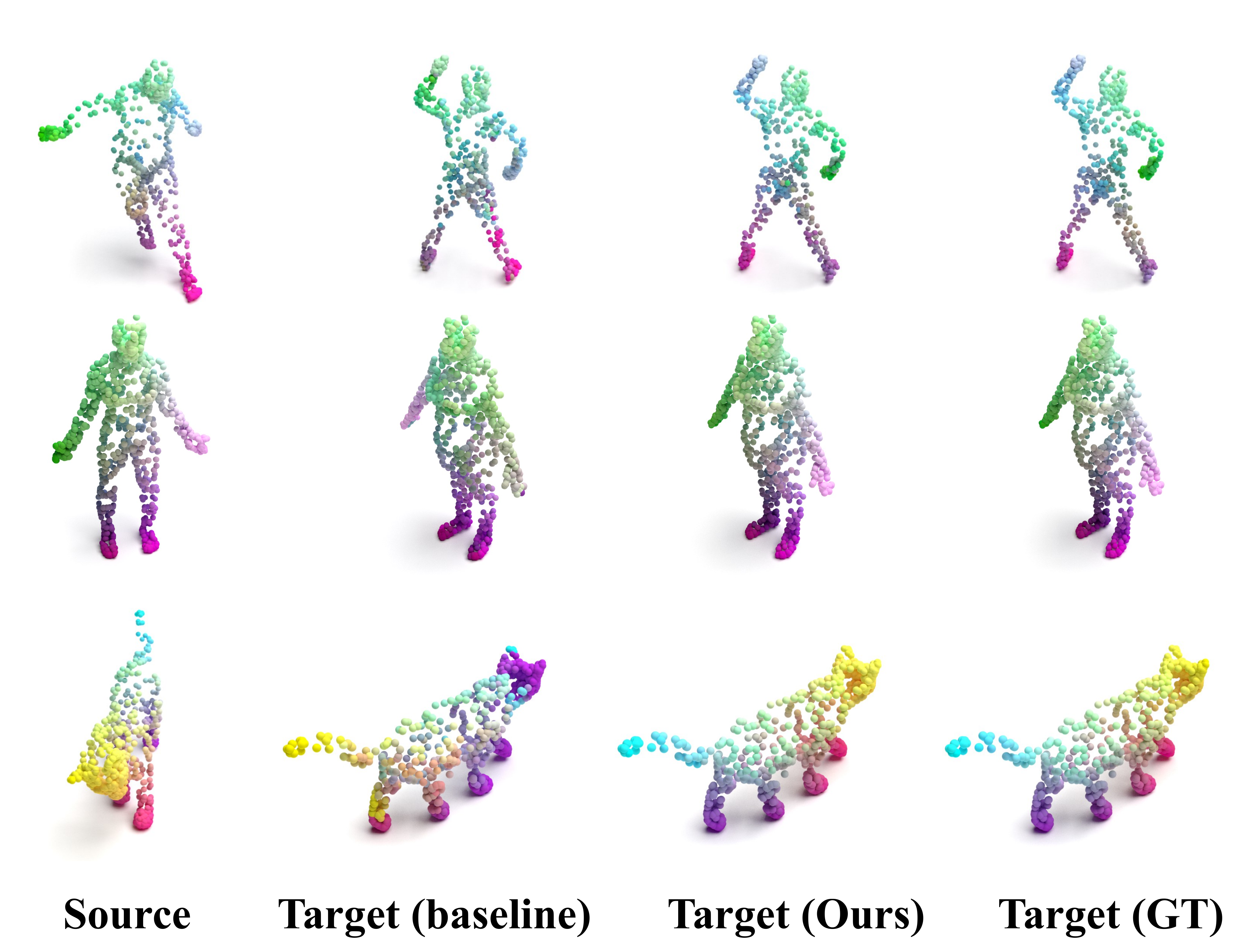}
    \vspace{-1em}
    \caption{
      \textbf{The visualization of dense point matching results. }
      Three point cloud pairs have different relative rotation angles. GT denotes ground truth. The correspondence is visualized by transferring colors from source to target according to matching results. The baseline predicts many false matches, especially for symmetrical, similar parts of the object. Our method achieves accurate matches for these parts with our orientation estimation module.}
    \label{fig:intro}
    \vspace{-2em}
  \end{figure}

To accurately find the point-to-point correspondence between two point clouds, spectral-based methods~\cite{bronstein2006generalized, huang2008non, tevs2011intrinsic, ovsjanikov2012functional, litman2013learning} have been proven as practical shape correspondence methods by computing functional mapping between the projected features and learning a transformation for the correspondence.
Nevertheless, the spectral-based methods suffer from complicated pre-processing steps and the necessity for connectivity information between points.
With the rapid development of deep learning, many fully supervised point cloud shape correspondence methods~\cite{deprelle2019learning, groueix20183d, marin2020correspondence} have been proposed to lead to remarkable progress. 
However, these methods rely on a large amount of carefully annotated point cloud pairs, which are expensive and time-consuming to collect.
%
%
%
To relieve the annotation cost of fully supervised methods, unsupervised methods~\cite{zeng2021corrnet3d, lang2021dpc} that utilize unlabeled data for model training have attracted more and more attention.
CorrNet3D~\cite{zeng2021corrnet3d} proposes the first unsupervised deep learning framework for building dense correspondence between point clouds in an end-to-end manner.
DPC~\cite{lang2021dpc} models the local point cloud structure by exploring the proximity of points using DGCNN~\cite{wang2019dynamic} and designs reconstruction losses to extract continuous point cloud representations.
However, in the scanning process of 3D scanner, due to light, vibration and other factors, point cloud noise will be inevitably generated.
Meanwhile, the pre-processing of point cloud (such as random subsampling) will also introduce noise.
Unfortunately, the previous methods fail to adequately consider the point cloud noise, which negatively impacts the point cloud representations. 
Besides the noise, existing methods lack attention to symmetrical parts of the body.
The mismatching issue of symmetrical parts is challenging in this task, which was also spotted by the previous method~\cite{zeng2021corrnet3d} but has yet to be solved.

By studying the previous point-based shape correspondence methods~\cite{deprelle2019learning, groueix20183d, marin2020correspondence, zeng2021corrnet3d, lang2021dpc}, we summarize two key issues that need consideration to achieve a more accurate shape correspondence:
1) \textit{How to overcome the disturbance of point cloud noise to get robust and consistent point cloud representations?}
Point cloud noise perturbs the spatial coordinates of point cloud and interferes with local structure modeling.
Therefore, it is necessary to overcome noise disturbances. 
2) \textit{How to solve the mismatching issue of symmetrical parts in point clouds with different body orientations?}
As shown in Figure~\ref{fig:intro}, for the pair of bilaterally symmetrical human point clouds facing the opposite directions, existing methods predict the completely reverse and seriously wrong point cloud correspondence due to the similar structure and position.
The specific relative rotation angles lead to severe mispredictions of symmetrical parts.
%
%
%

%
%

To achieve the above goals, we propose a \textit{Self-Ensembling Orientation-aware Network} (SE-ORNet)  for unsupervised point cloud shape correspondence.
We integrate orientation modeling and consistent point cloud representations under a unified self-ensembling framework, which consists of a pair of teacher and student models, an orientation estimation module, and an adaptive domain discriminator.
Firstly, we design a new augmentation scheme to produce augmented samples with rotation and Gaussian noise, and record the rotation angles as rotation angle labels. 
Then, we formulate soft labels and consistency losses to encourage consensus among ensemble predictions of augmented and raw samples, aiming to perceive the difference in body orientation and overcome the point cloud noise disturbance to obtain consistent point cloud representations.
%
%
In addition, we design a plug-and-play lightweight Orientation Estimation Module, which aligns the orientations of two point clouds to solve the mismatching issue of symmetrical parts in point clouds. 
%
%
Without  the real label of  relative rotation angle between the source and target,
we supervise the training with the rotation angle labels and calculate angle losses.
%
%
However, there is a noticeable domain gap between the rotation-augmented samples and the real samples. 
Therefore,
we design a discriminator to achieve domain adaptation and calculate the domain losses.
Furthermore, the discriminator facilitates the Orientation Estimation Module to mine the valuable knowledge in the rotation-augmented samples to compensate for the information loss of the real relative rotation angles.  

In summary, the main contributions of this work are as follows: (i) We design a plug-and-play lightweight Orientation Estimation Module that accurately aligns the orientations of point cloud pairs to achieve correct matching results of symmetrical parts.
(ii) We integrate point cloud orientation modeling and consistent point cloud representation learning with the disturbance of point cloud noise into a unified self-ensembling framework.
(iii) Our method attains state-of-the-art performance on both human and animal benchmarks, and extensive experimental results verify the superiority of our designs.

%% file: rw.tex
\begin{figure*}[!t]
    \vspace{-0.5em}
    \begin{center}
        \includegraphics[width=0.75\textwidth]{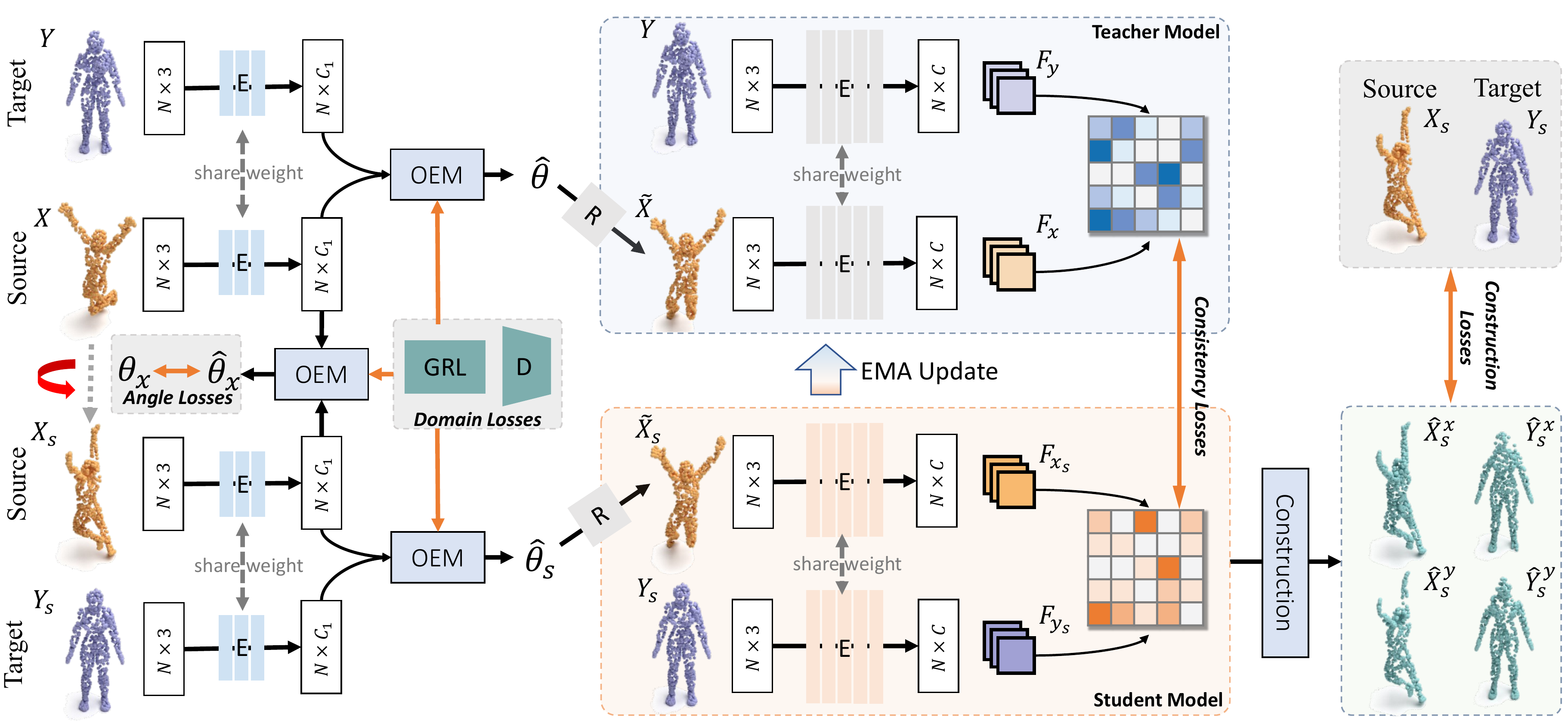}
        \vspace{-1em}
        \caption{\textbf{The overview of our self-ensembling orientation-aware network for unsupervised point cloud shape correspondence.}
        $X_s$ is generated from the raw source point cloud $X$ by random rotation and Gaussian noise, while $Y_s$ is only augmented by Gaussian noise.
        We design the Orientation Estimation Module to estimate the rotation $\theta$ of the source with respect to the target and align the point cloud in position space.
        Afterward, the aligned point cloud pairs are input to the teacher and student models, respectively, and the correspondence is predicted through a DGCNN backbone.
        Finally, we supervise the student model by the consistency losses and the construction losses, and the teacher model updates the parameters using the exponential moving average (EMA) strategy.
        The gradient reversal layer (GRL) acts as the identity function during forward propagation, but is multiplied by -1 during backward propagation.
        }
        \label{pipeline}
    \end{center}
    \vspace{-2.5em}
\end{figure*}

In this Section, we give a brief overview of related works on point cloud shape correspondence, including learning on point clouds, shape correspondence, and self-ensembling approaches.

\noindent\textbf{Learning on Point Clouds.}
%
%
%
%
%
%
PointNet~\cite{qi2017pointnet} learns from global information through multi-layer perceptrons and max-pooling operation.
PointNet++~\cite{qi2017pointnet++} devises a hierarchical architecture that recursively partitions the point cloud to extract local features more effectively.
Recent works explore local aggregators via relations~\cite{ran2021learning, yan2020pointasnl, zhao2021point}, and graphs~\cite{wang2019dynamic, zhou2021adaptive}.
PointCNN~\cite{li2018pointcnn} transforms neighboring points to the canonical order to apply traditional convolution on point clouds.
DGCNN~\cite{wang2019dynamic} creates a graph in the feature space and designs EdgeConv~\cite{wang2019dynamic} to learn the edge features of the graph in each layer.
However, the methods are commonly based on some assumptions of implicit local geometry, which may result in sensitivity to point cloud disturbances.

\noindent\textbf{Shape Correspondence.}
As an active research area in computer vision and graphics, point cloud shape correspondence methods roughly include spectral-based methods~\cite{bronstein2006generalized, huang2008non, tevs2011intrinsic, ovsjanikov2012functional, litman2013learning} and point-based methods~\cite{deprelle2019learning, groueix20183d, marin2020correspondence, zeng2021corrnet3d, lang2021dpc}.
Spectral-based methods require connectivity information to compute the LBO eigenvectors as basis functions and infer a linear transformation for shape correspondence.
However, with regard to point cloud data, connectivity information is difficult to obtain directly
while point-based methods directly process point clouds without connectivity information to find the dense point-to-point mapping between two point clouds with deformable 3D shapes.
Deprelle et al.~\cite{deprelle2019learning} propose representing shapes as the deformation and combination of learnable elementary 3D structures.
Groueix et al.~\cite{groueix20183d} employ an encoder-decoder architecture to obtain and constrain the similarity matrix with manually annotated labels.
The deep learning methods train their neural networks in a data-driven manner and improve performance to a large extent.
However, manually labeling the point-to-point correspondence between two point clouds in 3D space takes much time and effort.
Therefore, some unsupervised point cloud shape correspondence methods~\cite{zeng2021corrnet3d, lang2021dpc} are proposed to reduce the overhead of labeling.
CorrNet3D~\cite{zeng2021corrnet3d} is the first unsupervised deep learning framework.
DPC~\cite{lang2021dpc} designs several reconstruction losses to smooth point cloud representations. 
Due to the lack of annotation, the mismatching issue of symmetrical parts in point clouds with different orientations has become an undeniable problem in unsupervised shape correspondence area. 

\noindent\textbf{Self-ensembling Approaches.}
Self-ensembling approaches improve the model generalization by encouraging consensus among ensemble predictions of unknown samples with small perturbations. 
$\Gamma$ model~\cite{rasmus2015semi} consists of two identical parallel branches that respectively take raw images and corrupted images as inputs. 
In contrast to $\Gamma$ model, $\Pi$ model~\cite{laine2016temporal} integrates two parallel branches into a single branch.
As an extension of the $\Pi$ model, the temporal model~\cite{laine2016temporal} forces the consistency between the outputs and the aggregated outputs over previous training epochs.
Mean Teacher~\cite{tarvainen2017mean} replaces network prediction average with network parameter average via the exponential moving average (EMA) strategy. 
We design a framework similar to Mean Teacher and adapt it for the unsupervised point cloud shape correspondence task.
The proposed framework facilitates the network to yield consistent and accurate predictions under noise perturbations and orientation rotations.

%% file: method.tex
\subsection{Overview}
\label{Overview}
The unsupervised point cloud shape correspondence is to find the mapping $f: X \rightarrow Y$ between two point clouds (source $X$ and target $Y$) without ground-truth correspondence annotations.
Figure~\ref{pipeline}\ illustrates the pipeline of our approach.
Given source $X$ and target $Y$ point clouds, we utilize random rotation and Gaussian noise to generate the augmented source point cloud $X_s$, while we augment $Y$ by Gaussian noise to obtain $Y_s$.
Due to the absence of the relative rotation angle labels $\theta, \theta_s$, we use the rotation angle labels $\theta_x$ to guide the Orientation Estimation Module learning and transfer  the valuable information by the adversarial domain adaptation method.
The aligned point cloud pairs are input to the teacher and student models, respectively,
and we use the DGCNN~\cite{wang2019dynamic} backbone to output the similarity matrices.
After that, we generate two reconstructed point clouds $\hat{X}, \hat{Y}$ based on the predicted similarity matrices by the student model.
Finally, the student model is supervised by the consistency losses and the construction losses, while the parameters of teacher model are updated by the exponential moving average (EMA) strategy.

\subsection{Orientation Estimation}
\label{Domain_Adaptation}

\begin{figure}[!t]
    \centering
    \includegraphics[width=0.47\textwidth]{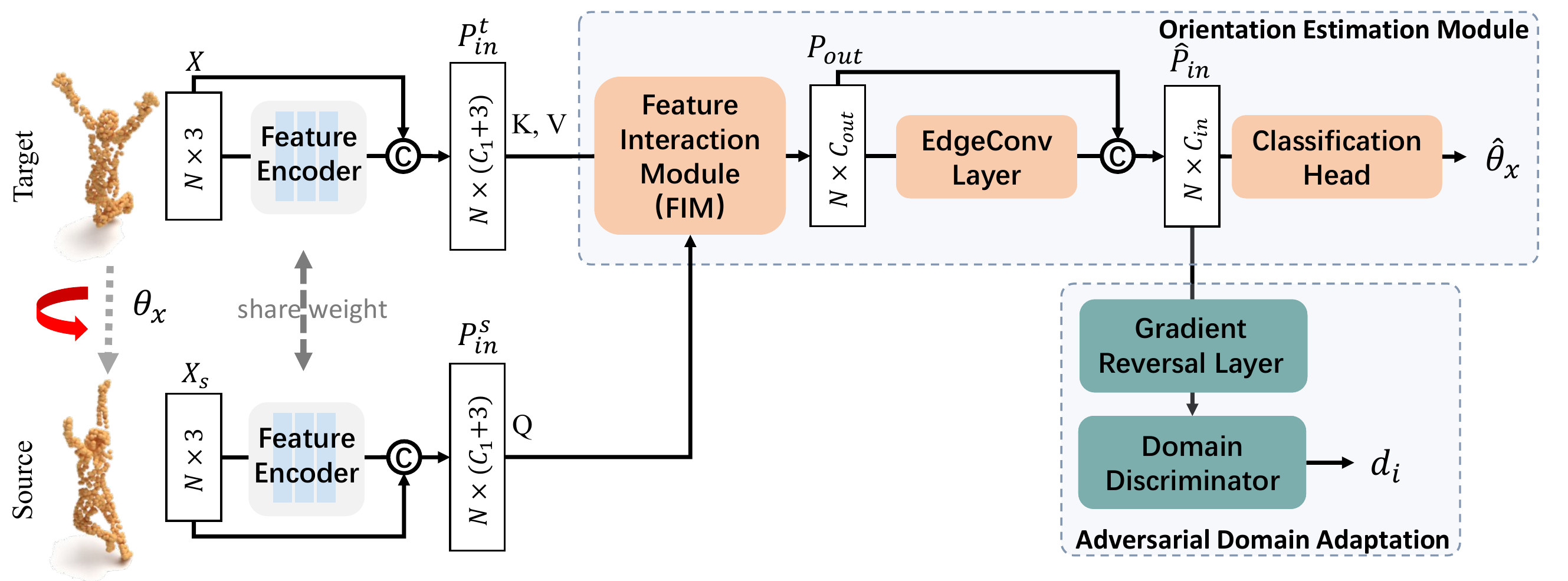}
    \scriptsize
    \vspace{-2em}
    \caption{\textbf{The overview of Orientation Estimation.}}
    \label{fig:overview_oem}
    \vspace{-2em}
  \end{figure}

We provide the overview of the orientation estimation in Figure~\ref{fig:overview_oem}. The encoded features $P_{in}^s, P_{in}^t$ are fed into the Feature Interaction Module (FIM) for feature fusion. The fused features $P_{out}$ are enhanced by a single-layer edgeconv. To predict the relative orientation, the features $\hat{P}_{in}$ are fed into the classification head. We also input $\hat{P}_{in}$ into the discriminator to determine whether the pair comes from the same shape or from different shapes.

\begin{figure}[!t]
    \begin{center}
        \includegraphics[width=0.28\textwidth]{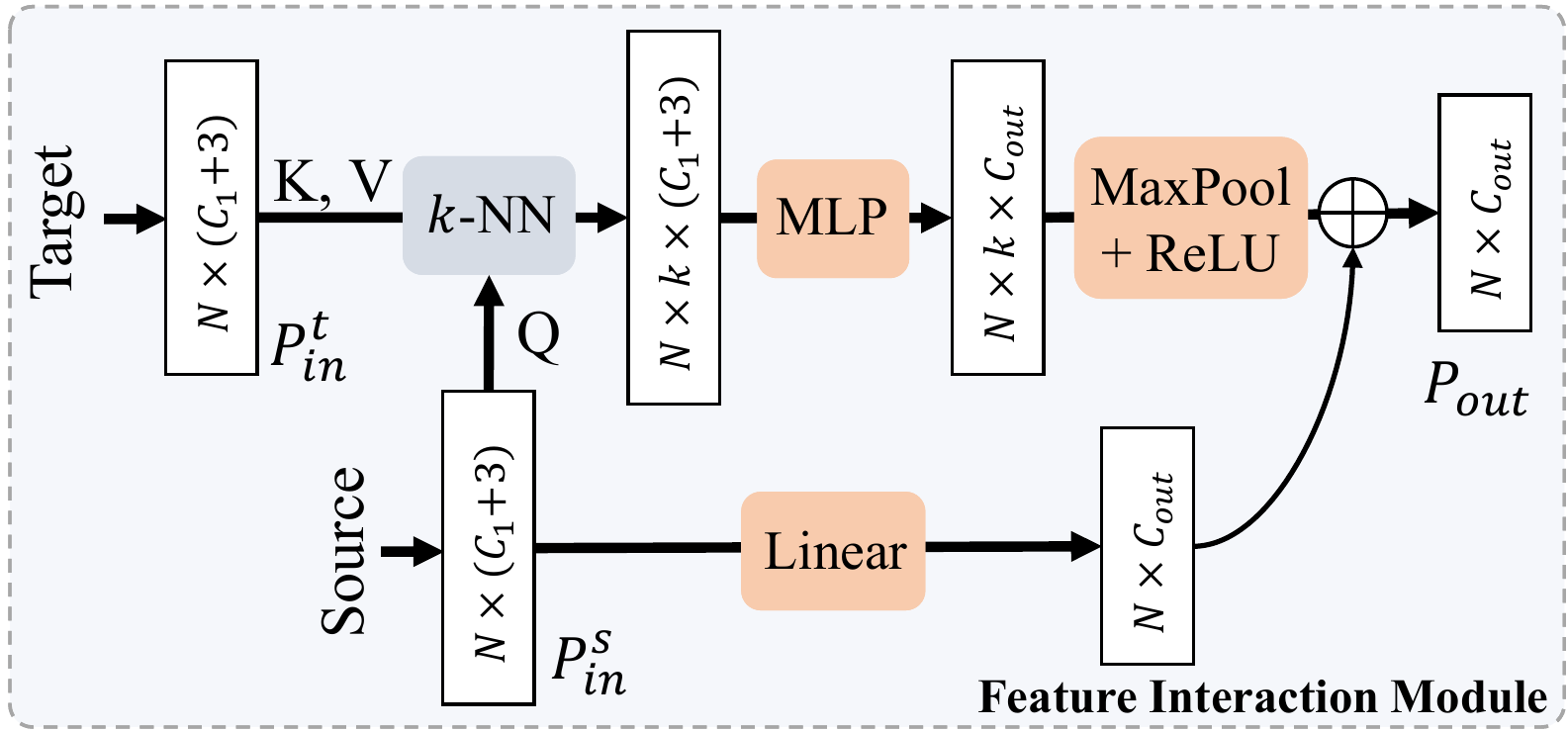}
        \vspace{-1em}
        \caption{\textbf{The Feature Interaction Module (FIM).}}
        \vspace{-2.5em}
        \label{interaction}
    \end{center}
\end{figure}

\textbf{Feature Interaction Module.}
%
As shown in Figure~\ref{interaction}, the feature interaction module is a query-based graph convolution.
The point features in the source point cloud are updated by querying points with similar features in the target point cloud.
%
Let $P^s_{in}, P^t_{in}$ be the inputs of the feature interaction module.
For each point $p_i \in P^s_{in}$, we first consider it as a query to search for the k-nearest neighbors ${q^j_i, 1\leq j \leq k}$ in the target point cloud $P^t_{in}$ with respect to the Euclidean distance defined in the $C_1$-dimensional feature space.
To better model the relative rotation relationships in space, we use spatial position differences and feature differences as features of each edge, denoted as $(p_i, q^j_i - p_i)$, where $p_i, q^j_i \in \mathcal{R}^{(C_1 + 3)}$.
Then we use a multilayer perceptron (MLP) to compute a new feature $e_{ji} = \operatorname{MLP}(p_i, q_i^j - p_i)$ from each edge.
For each point, we aggregate its k edge features into a new point feature through Max pooling and ReLU activation.
In addition, we add a linear layer that skip-connects the output $P_{out}$ with the input to make the block residual.

\textbf{Rotation Classification Head.}
We concatenate the output feature $P_{out}$ and the enhanced feature as the input $\hat{P}_{in}$ of the Rotation Classification Head.
Inspired by the orientation prediction in the point cloud detection methods~\cite{qi2018frustum, mousavian20173d}, we consider the prediction of relative rotation angle as a classification task.
That coarsely aligning the orientations of the source and target point clouds is enough to solve the problem of mismatching issue of symmetrical parts in point clouds.
Thus, we pre-define $M$ equally divided angle bins and then use an MLP head to classify the relative rotation angle into those pre-defined categories.
Specifically, we compress $\hat{P}_{in}$ using maximum pooling and average pooling to obtain the global features and
then predict the probability distribution of the relative rotation angle.
Finally, we choose the angle bin with the highest probability as the relative rotation angle of the source and target point clouds.

\textbf{Adversarial Domain Adaptation.}
Due to the absence of relative rotation angle $\theta, \theta_s$,
we utilize the relative rotation angle $\theta_x$ to guide the Orientation Estimation Module learning.
However, there is a noticeable domain gap between the rotation-augmented samples and the real samples.
To eliminate the domain gap, we use a discriminator to identify whether the input features $\hat{P}_{in}$ of the classification head are from the rotation-augmented samples or the real samples.
Specifically, we use a PointNet-like module to process the features $\hat{P}_{in}$ and predict the probability $d_i$ that the $P_{in}$ is from the real samples:
\begin{equation}
    \label{PointNet}
    d_{i}=\operatorname{MLP}_{2}\left\{\max _{N}\left\{\operatorname{MLP}_{1}\left[\hat{P}_{in}\right]\right\}\right\},
\end{equation}
where $\max$ refers to the global Max pooling. 
By using the discriminator, we train the model to disregard the distinction between the two domains and instead prioritize learning the relative rotation information.
%

\textbf{Relative Rotation Angle Supervision.}
The angle loss is computed with a cross entropy loss:
\begin{equation}
    \label{cross_entropy}
    \mathcal{L}_{angle} = -\sum_{i=1}^{M} \theta^i_x \log \left(\hat{\theta}^i_x\right),
\end{equation}
where $M$ is the number of the pre-defined angle bins, $\hat{\theta}^i_x$ and $\theta^i_x$ denote the predicted probability distribution of relative rotation angle and the rotation angle label, respectively.
The domain loss is computed with a Focal Loss~\cite{lin2017focal}:
\begin{equation}
    \label{Focal_loss}
    \mathcal{L}_{domain}=-\left(1-d_{i}\right)^{\gamma} \log \left(d_{i}\right), \gamma>1,
\end{equation}
where $\mathcal{L}_{domain}$ denotes the domain loss and $\gamma$ denotes the tunable focusing parameter.

\subsection{Self-Ensembling Framework}
\label{Ensemble_Framework}
As shown in Table~\ref{table:robustness}, the point cloud orientation and Gaussian noise are two crucial factors in the unsupervised point cloud shape correspondence task.
The existing methods ignore noise interference on the point cloud correspondence, making it difficult to obtain robust point cloud representations.
To address the above problems, we utilize the Mean Teacher architecture~\cite{tarvainen2017mean} and design two consistency losses to constrain the student model for robust feature representations under orientation disturbance and Gaussian noise interference.

\textbf{Stochastic Transform.}
We apply stochastic transformations that include rotation and Gaussian noise on the point clouds for the student network formulated as $\tau = (\mathcal{R},\mathcal{N})$.
More specifically, given the raw point cloud pair $(X,Y)$, we first apply  Gaussian noise to the target point cloud $Y$:
\begin{equation}
    \label{Gaussian_noise}
    y^s_i=y_i + n_i, n_i \sim \mathcal{N}(0, \sigma^2),
\end{equation}
where $\mathcal{N}(0, \sigma^2)$ means the Gaussian distribution with the 0 mean and $\sigma$ standard deviation.
Then, we utilize Gaussian noise and random rotation along the vertical z-axis to augment the source point cloud $X$:
\begin{equation}
    \label{random_rotation}
    x^s_i= \mathcal{R}(\theta_x) \odot x_i + n_i, n_i \sim \mathcal{N}(0, \sigma^2),
\end{equation}
where $\mathcal{R}$ denotes the rotation around the z-axis and $\theta_x$ is the rotation angle sampled from $[0, 2\pi]$.

\textbf{Architecture of Teacher \& Student Models.}
Given the raw and augmented point cloud pairs $(X,Y), (X_s,Y_s)$,
we use the  Orientation Estimation Module to align source and target orientations.
Then, our approach follows the Mean Teacher paradigm~\cite{tarvainen2017mean} and inputs the aligned point cloud pairs $(\tilde{X}_s,Y_s), (\tilde{X},Y)$ into the student and teacher models, respectively.
We use a variant of DGCNN~\cite{wang2019dynamic} backbone to embed the aligned source $\tilde{X}$ and target $Y$ point clouds to a high-dimensional feature space as $F_x, F_y \in \mathcal{R}^{N \times C}$.
Unlike DGCNN, which uses dynamic graphs to select neighbors, we follow \cite{lang2021dpc} to use static graphs for the neighbors.
Specifically, we select the k-nearest neighbors in the  coordinate space instead of the feature space.
Then, we use the cosine similarity $S_{xy}$ of   source and target point features $F_x, F_y$  to measure their correspondence.
\begin{equation}
    \label{cosine_similarity}
    s_{i j}=\frac{F_{x}^{i} \cdot (F_{y}^{j})^{T}}{\|F_{x}^{i}\|_{2} \|F_{y}^{j}\|_{2}},
\end{equation}
where $F_{x}^{i}, F_{y}^{j}$ are the i$^{\text{th}}$ and j$^{\text{th}}$ rows of $F_x, F_y$, respectively, and $(\cdot)^T$ denotes a transpose operation.

\textbf{Soft Label \& Consistency Loss.}
To enhance the perception of local structures and the robustness of the model, we design two consistency losses to maintain feature consistency under Gaussian noise and orientation disturbances.
We utilize the cross-similarity $S^t_{xy}$ obtained from the teacher model as a soft label and the $\operatorname{Smooth} L_1$ loss to constrain the cross-similarities $S^s_{x_sy_s}$ of the student model:
\begin{equation}
    \label{cross_consistancy}
    \mathcal{L}_{ccs} = \operatorname{Smooth} L_1(S^t_{xy}, S^s_{x_sy_s}),
\end{equation}
where $\mathcal{L}_{ccs}$ denotes the consistency loss of the cross-similarities.
To reduce the interference of point cloud orientation on correspondence estimation, we model the feature consistency of the strongly augmented source $X_s$ and raw source $X$.
We constrain the consistency of the self-similarities $S_{xx}$ to ensure the consistency of neighboring points with similar features.
We use the feature embedding $F_x$ of the source point cloud to compute the self-similarity $S_{xx}$ by Equation~\eqref{cosine_similarity}.
Then, we use the $\operatorname{Smooth} L_1$ loss to constrain the self-similarities $S^s_{x_sx_s}$ of the student model with the self-similarities $S^t_{xx}$ from the teacher model:
\begin{equation}
    \label{self_consistancy}
    \mathcal{L}_{css} = \operatorname{Smooth} L_1(S^t_{x x}, S^s_{x_s x_s}),
\end{equation}
where $\mathcal{L}_{css}$ denotes the consistency loss of the self-similarities.
The above two consistency losses ensure the robustness of feature embedding under Gaussian noise and orientation disturbances for accurate correspondence.

\subsection{Model Training \& Inference}
\label{Model_Training}
In addition to the above mentioned consistency losses, angle loss, and  domain loss,
we use reconstruction losses to promote a unique point matching between the shape pair.
%
Following the previous work~\cite{lang2021dpc}, we perform the cross-construction operation to construct the target shape $\hat{Y}$ by using the feature similarity $S_{xy}$ between source and target point clouds, and the target point coordinates ${Y}$ as follows:
\begin{equation}
    \label{construction}
    \hat{y}_{x_{i}}=\sum_{j \in \mathcal{N}_{\mathcal{Y}}\left(x_{i}\right)} \frac{e^{s_{i j}}}{\sum_{l \in \mathcal{N}_{\mathcal{Y}}(x_{i})} {e^{s_{i l}}}} y_{j},
\end{equation}
where $\mathcal{N}_{\mathcal{Y}}(x_{i})$ is latent k-nearest neighbors of $x_{i}$ in the target $Y$.
When the source and target point clouds are identical, we refer to the construction operation as self-construction.
As shown in Figure~\ref{pipeline}, we obtain the point clouds $\hat{Y}^x_s, \hat{X}^x_s, \hat{X}^y_s, \hat{Y}^y_s$ by cross-construction and self-construction.
Then, we constrain the training with the construction loss as follows:
\begin{equation}
\begin{aligned}
    \label{construction_loss}
    \mathcal{L}_{cons}=&\lambda_{cc}(\operatorname{CD}(Y_s, \hat{Y}^x_s) + \operatorname{CD}(X_s, \hat{X}^y_s)) \\
    &+ \lambda_{sc}(\operatorname{CD}(Y_s, \hat{Y}^y_s) + \operatorname{CD}(X_s, \hat{X}^x_s)),
\end{aligned}
\end{equation}
where $\lambda_{cc}, \lambda_{sc}$ are hyperparameters and
$\operatorname{CD}$ means the Chamfer Distance.
Finally, we add a regularization term to correspond close points in the source to close points in the target.
\begin{equation}
    \label{regularization}
    \mathcal{L}_{norm}=\sum_{i} \sum_{l \in \mathcal{N}_{\mathcal{Y}}\left(x_{i}\right)} e^{\left\|x_{i}-x_{l}\right\|_{2}^{2} / \alpha} \left\|\hat{y}_{x_{i}}-\hat{y}_{x_{l}}\right\|_{2}^{2},
\end{equation}
where $\mathcal{N}_{\mathcal{Y}}(x_{i})$ is the same as defined in Equation~\ref{construction}
and $\alpha$ is a hyperparameter.
To sum up, the total loss of our unsupervised point cloud shape correspondence method is:
\begin{equation}
\begin{aligned}
    \label{overall_loss}
    \mathcal{L}_{total}= &\lambda_1 \mathcal{L}_{ccs} + \lambda_2 \mathcal{L}_{css} + \lambda_3 \mathcal{L}_{angle} \\ & + \lambda_4 \mathcal{L}_{domain} + \mathcal{L}_{cons} + \lambda_5 \mathcal{L}_{norm},
\end{aligned}
\end{equation}
where $\lambda_i$ is a hyperparameter, balancing the contribution of   different loss terms.
%
During inference, we set the closest point $y_j^{*}$ in the feature space for each point $x_i$ as its corresponding point.
This selection rule can be formulated as:
\begin{equation}
        \label{inference}
        f\left(x_{i}\right)=y_{j^{*}}, j^{*}=\underset{j}{\operatorname{argmax}} (s_{i j}).
\end{equation}

%% file: expr.tex
\subsection{Experimental Setup}
\textbf{Dataset.}
To demonstrate the effectiveness and generalization of our method, we perform experiments on human and animal datasets.
We conduct experiments on human datasets according to DPC's~\cite{lang2021dpc} scheme.
For the large-scale dataset, we randomly downsample the SURREAL~\cite{groueix20183d} dataset, which contains 230000 training shapes, into 2000 shape pairs as the training set.
For the test set, we use the SHREC'19~\cite{melzi2019shrec} dataset, which contains 44 real human models, and pair them into 430 annotated test examples.
To further verify the ability of our method to learn discriminative feature expression with a small data size, we train SE-ORNet on the pairs randomly sampled from 44 SHREC instances, and the testing is still conducted on the official 430 SHREC'19 pairs.

For animal datasets, we also conduct experiments with different dataset scales.
We use the large-scale SMAL~\cite{zuffi20173d} dataset and TOSCA~\cite{bronstein2008numerical} dataset as the training set and test set, respectively.
SMAL dataset consists of parameterized models of various animals, and we randomly sample SMAL to obtain the corresponding shape pairs as the training set.
TOSCA is generated by deforming three template meshes (human, dog, and horse) into multiple poses.
We pair the 41 animal figures in TOSCA from the same category to
%
form a training set of 260 samples and a test set of 286 samples. 
Because the number of points in different shapes varies, we make a random downsample of the original point cloud to a fixed number $n=1024$ , as done in  CorrNet3D~\cite{zeng2021corrnet3d}.

\textbf{Evaluation Metrics.}
The evaluation metrics include the average correspondence error and the correspondence accuracy.
Based on the Euclidean-based measure, the average correspondence error is defined for a pair of source and target shapes $(X,Y)$ as follows:
\begin{equation}
	\label{correspondence error}
	err=\frac{1}{n} \sum_{x_{i} \in X}\left\|f\left(x_{i}\right)-y_{g t}\right\|_{2},
\end{equation}
where $y_{g t} \in Y$ is the ground-truth matching point to $x_{i}$. 
The unit is centimeter(cm).
And the correspondence accuracy can be formulated as:
\begin{equation}
	\label{correspondence accuracy}
	acc(\epsilon)=\frac{1}{n} \sum_{x_{i} \in X} \mathbb{I}\left(\left\|f\left(x_{i}\right)-y_{g t}\right\|_{2}<\epsilon d\right),
\end{equation}
where $\mathbb{I} (\cdot)$ is the indicator function, $d$ is the maximal Euclidean distance between points in $Y$, and $\epsilon \in [0, 1]$ is an error tolerance. 

\textbf{Implementation Details.}
For a fair comparison with existing methods~\cite{lang2021dpc, zeng2021corrnet3d},
we use the same DGCNN~\cite{wang2019dynamic} backbone with
four EdgeConv blocks as the feature extractor in the self-ensembling framework.
The standard deviation $\sigma$ in Equation~\eqref{Gaussian_noise} is set as 0.1 for human datasets and 0.15 for animal datasets.
In the Orientation Estimation Module,
the feature encoding module is a simplified DGCNN with
three EdgeConv~\cite{wang2019dynamic} blocks whose layer
output sizes are 64, 128, and 256.
The $k$ of k-NN is set as 24,
and the slope of all LeakyReLU is 0.2.
We feed the output of the last layer into
the proposed feature interaction module
with the output size 256.
Then we refine the feature by an EdgeConv layer with the same output size.
%
%
The angle classification head consists of three Linear-BN-ReLU and an output Linear. 
The channels are 256, 128, and 128 for the three Linear-BN-ReLU.
The last Linear outputs the probability for classification.
%
%
We set the number of bins $M$ as 8 in the angle classification head, where each bin represents a range of $45^\circ$.
The domain discriminator is a PointNet-like module consisting of two MLPs in Equation~\eqref{PointNet}. 
The channels of $\operatorname{MLP_1}$ are 512, 256, and 128, while the channels of $\operatorname{MLP_2}$ are 256, 128, and 256.
%
We follow~\cite{lang2021dpc} and use a neighborhood size $k = 10$ in Equation~\eqref{construction} and~\eqref{regularization}. 
$\lambda_cc$ and $\lambda_sc$ in Equation~\eqref{construction_loss} are set as 1 and 10, respectively. $\lambda_1$, $\lambda_2$, $\lambda_3$, $\lambda_4$, and $\lambda_5$ in Equation~\eqref{overall_loss} are set as 0.1, 0.1, 1.0, 0.8, and 1.0, respectively.

\subsection{Comparison on Human Datasets}
For a fair comparison with existing methods, we do not use any post-processing or additional connectivity information.
%
%
In addition, we follow DPC~\cite{lang2021dpc} and train our model on the SURREAL and SHREC datasets, respectively. Then we test our model on the official 430 SHREC'19 pairs. 
%


\textbf{Evaluation on SHREC dataset.}
\begin{table}[!t]
  \begin{center}
    \footnotesize
    \setlength\tabcolsep{6pt}
    \caption{\textbf{Comparison on SHREC and SURREAL benchmarks.} Here, acc means the correspondence accuracy at an error tolerance of 0.01, while err refers to the average correspondence error. Higher accuracy and lower error reflect a better result. 
    }
    \label{table:Human_dataset}
    \vspace{-0.8em}
    \begin{tabular}{c|c|cc|cc}
      \toprule
      \multirow{2}*{Method} & \multirow{2}*{Input} & \multicolumn{2}{c|}{SHREC} & \multicolumn{2}{c}{SURREAL} \\
      \cline{3-4}\cline{5-6}
      & & acc $\uparrow$ & err $\downarrow$ & acc $\uparrow$ & err $\downarrow$ \\
      \midrule
      Diff-FMaps\cite{marin2020correspondence}  & Point & / & / & 4.0\% & 7.1 \\
      3D-CODED\cite{groueix20183d}              & Point & / & / & 2.1\% & 8.1 \\
      Elementary\cite{deprelle2019learning}     & Point & / & / & 2.3\% & 7.6 \\
      CorrNet3D\cite{zeng2021corrnet3d}         & Point & 0.4\%  & 33.8 & 6.0\% & 6.9  \\
      DPC\cite{lang2021dpc}                     & Point & 15.3\%  & 5.6 & 17.7\% & 6.1 \\
      \textbf{Ours}                             & Point & \textbf{17.5\%}  & \textbf{5.1} & \textbf{21.5\%} & \textbf{4.6} \\
      \bottomrule
    \end{tabular}
    \vspace{-2em}
  \end{center}
\end{table}
\begin{figure}[!t]
  \begin{center}
      \includegraphics[width=0.37\textwidth]{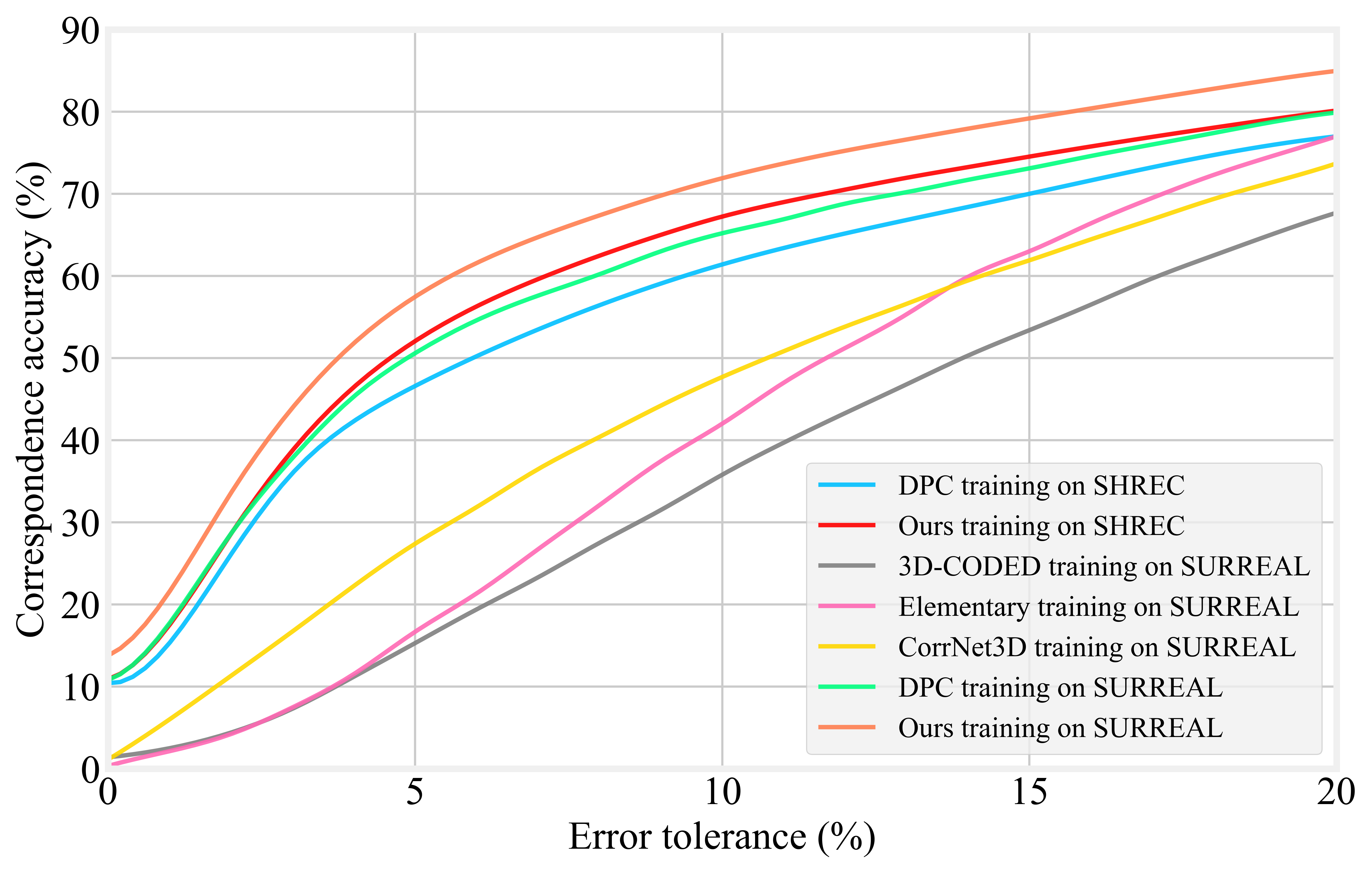}
      \vspace{-1em}
      \caption{\textbf{Correspondence accuracy at various error tolerances for human datasets.} The methods are trained on the SHREC or SURREAL dataset and evaluated on SHREC'19 test pairs. Compared with other methods, our approach achieves an impressive performance improvement.}
      \vspace{-2.5em}
      \label{human}
  \end{center}
\end{figure}
As shown in Table~\ref{table:Human_dataset}, our approach shows significant performance improvements
on the SHREC benchmark and achieves new SOTA performance by 2.2\% improvements in accuracy and 0.5 reductions in error.
To show the improvement under different error tolerances, we present the correspondence accuracy for point-based methods trained on SHREC and evaluated on the SHREC'19 test set.
As shown in Figure~\ref{human}, our method achieves better results with different error tolerances. 

\textbf{Cross-dataset Generalization.}
In Table~\ref{table:Human_dataset} and Figure~\ref{human}, we also report the comparison with other methods on the SURREAL benchmark.
The models are trained on the SURREAL dataset and evaluated on the SHREC'19 test set.
The large-scale training set of the SURREAL dataset helps the deep learning-based methods perform better, even though there is a domain gap between the SURREAL and SHREC datasets.
With our proposed method, the correspondence accuracy reaches 21.5\% at an error tolerance of 0.01, and the average correspondence error is reduced to 4.6 on the SHREC'19 test set.

\subsection{Comparison on Animal Datasets}
\begin{figure}[!t]
  \begin{center}
      \includegraphics[width=0.37\textwidth]{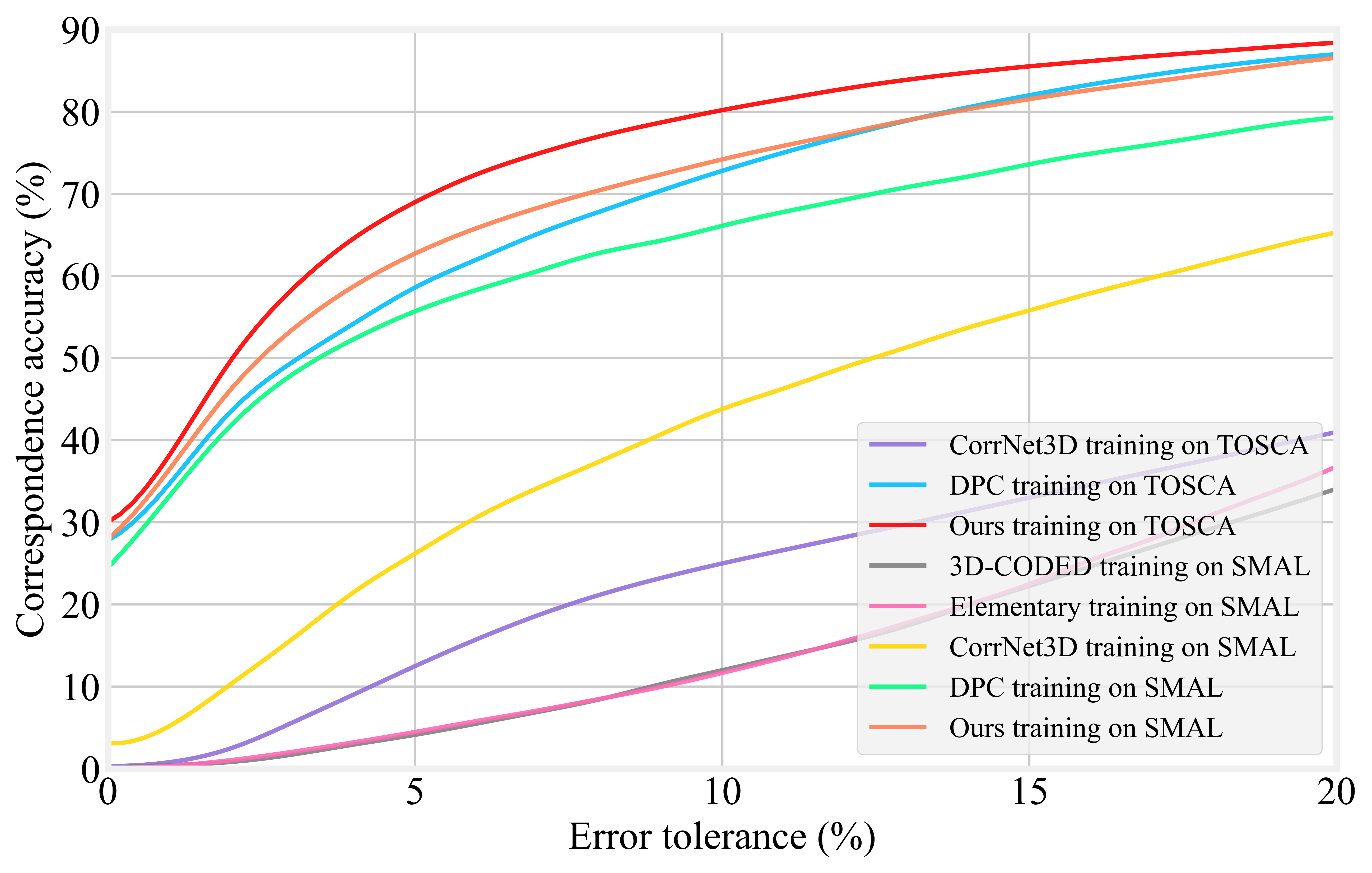}
      \vspace{-1em}
      \caption{\textbf{Correspondence accuracy at various error tolerances for animal datasets.} The methods are trained on the TOSCA or SMAL dataset and evaluated on the official TOSCA test pairs. Compared with other methods,  our method achieves a desirable performance improvement. 
      }
      \vspace{-2.5em}
      \label{animal}
  \end{center}
\end{figure}

To verify the adaptability of our method to point clouds of different shapes, we conduct experiments on two animal benchmarks.
Similar to human datasets, we train our method on TOSCA and SMAL datasets respectively, and test on the TOSCA test dataset.
Table~\ref{table:Animal_dataset} and Figure~\ref{animal} show the competitive results on TOSCA and SMAL benchmarks.
Our method achieves a 3.5\% accuracy improvement on the TOSCA benchmark and a 3.2\% accuracy improvement on the SMAL benchmark.
Compared with the human datasets, the animal datasets have various shapes with aligned orientations.
Thus, the above performance gains come mainly from our self-ensembling framework, which can learn reliable features on complex data.
To verify the effect of the Orientation Estimation Module, we test our method and baseline using different augmented test sets in Section~\ref{Robustness_Analysis}.

\begin{table}[!t]
  \begin{center}
    \footnotesize
    \setlength\tabcolsep{6pt}
    \caption{\textbf{Comparison on TOSCA and SMAL benchmarks.} Here, acc means the correspondence accuracy at an error tolerance of 0.01, while err refers to the average correspondence error. Higher accuracy and lower error reflect a better result.}
    \label{table:Animal_dataset}
    \vspace{-0.8em}
    \begin{tabular}{c|cc|cc}
      \toprule
      \multirow{2}*{Method}  & \multicolumn{2}{c|}{TOSCA} & \multicolumn{2}{c}{SMAL} \\
      \cline{2-3}\cline{4-5}
      & acc $\uparrow$ & err $\downarrow$ & acc $\uparrow$ & err $\downarrow$ \\
      \midrule
      3D-CODED\cite{groueix20183d}               & / & / & 0.5\% & 19.2 \\
      Elementary\cite{deprelle2019learning}      & / & / & 0.5\% & 13.7 \\
      CorrNet3D\cite{zeng2021corrnet3d}          & 0.3\%  & 32.7 & 5.3\% & 9.8  \\
      DPC\cite{lang2021dpc}                      & 34.7\%  & 2.8 & 33.2\% & 5.8 \\
      \textbf{Ours}                              & \textbf{38.2\%}  & \textbf{2.7} & \textbf{36.4\%} & \textbf{3.9} \\
      \bottomrule
    \end{tabular}
    \vspace{-2.5em}
  \end{center}
\end{table}


\subsection{Comparison on Real-world Dataset}
CMU Panoptic~\cite{joo2015panoptic} is a dataset of scanned point clouds of human subjects in various poses, containing noise, outliers, occlusions, and clutter. 
Meanwhile, SHREC’20~\cite{dyke2020shrec} dataset contains real scans of various four-legged animal models. 
As shown in Table~\ref{table:panoptic} and Table~\ref{table:shrec20}, we provide the results under two real scan datasets, demonstrating the remarkable robustness of our model to noise.

\begin{table}[!t]
  \begin{center}
    \footnotesize
    \setlength\tabcolsep{2.5pt}
    \caption{\textbf{Comparison on CMU-Panoptic benchmark.} Here, err means average Euclidean keypoint error (cm). }
    \label{table:panoptic}
    \vspace{-1.3em}
    \begin{tabular}{c|ccccc}
      ~  & 3D-CODED~\cite{groueix20183d} & DIF-Net~\cite{ren2018continuous} & CorrNet~\cite{zeng2021corrnet3d}  & IFMatch~\cite{sundararaman2022implicit} & \textbf{Ours} \\ \hline
      err & 17.1     & 15.3    & 14.8     & 8.5                                    & \textbf{3.2}  \\
    \end{tabular}
    \vspace{-1.5em}
  \end{center}
\end{table}

\begin{table}[!t]
  \begin{center}
    \footnotesize
    \setlength\tabcolsep{8pt}
    \vspace{-1.1em}
    \caption{
        \textbf{Comparison on SHREC'20 benchmark.} The training dataset is indicated in the bracket.
    }
    \label{table:shrec20}
    \vspace{-1.5em}
    \begin{tabular}{c|cc|cc}
      \multirow{2}*{Method} & \multicolumn{2}{c|}{SHREC'20 [SURREAL]} & \multicolumn{2}{c}{SHREC'20 [SMAL]}                                      \\
      \cline{2-3}\cline{4-5}
                            & acc $\uparrow$                          & err $\downarrow$                    & acc $\uparrow$  & err $\downarrow$ \\
      \hline
      DPC\cite{lang2021dpc}                   & 25.0\%                                  & 3.2                                 & 24.5\%          & 7.5              \\
      \textbf{Ours}         & \textbf{29.9\%}                         & \textbf{1.2}                        & \textbf{25.4\%} & \textbf{2.9}     \\
    \end{tabular}
    \vspace{-2.5em}
  \end{center}
\end{table}


\subsection{Ablation Study}

\begin{table}[!t]
  \begin{center}
    \footnotesize
    \setlength\tabcolsep{5pt}
    \centering
    \caption{\textbf{Evaluation of the model with different designs on SURREAL.} $\mathcal{L}_{css}$ is the consistency loss of the self-similarities, $\mathcal{L}_{ccs}$ is the consistency loss of the cross-similarities, $\tau$ means the stochastic transform, OEM means we use the Orientation Estimation Module, FIM is the Feature Interaction Module, and DAM is the Domain Adaptation Module.}
    \vspace{-0.8em}
    \begin{tabular}{cccccc|cc}
      \toprule
      \multirow{2}*{$\tau$} &\multirow{2}*{$\mathcal{L}_{ccs}$} &  \multirow{2}*{$\mathcal{L}_{css}$} &   \multirow{2}*{OEM} & \multirow{2}*{FIM} & \multirow{2}*{DAM} & \multicolumn{2}{c}{SURREAL} \\
      \cline{7-8}
      & & & & & & acc $\uparrow$ & err $\downarrow$ \\
      \midrule
      \ding{55}&\ding{55}&\ding{55}&\ding{55}&\ding{55}&\ding{55}&17.7\%&6.1\\
      \ding{51}&\ding{51}&\ding{55}&\ding{55}&\ding{55}&\ding{55}&18.8\%&5.7\\
      \ding{51}&\ding{51}&\ding{51}&\ding{55}&\ding{55}&\ding{55}&19.2\%&5.6\\
      \ding{51}&\ding{51}&\ding{51}&\ding{51}&\ding{55}&\ding{55}&19.5\%&5.5\\
      \ding{51}&\ding{51}&\ding{51}&\ding{51}&\ding{51}&\ding{55}&20.4\%&5.1\\
      \ding{51}&\ding{51}&\ding{51}&\ding{51}&\ding{51}&\ding{51}&\textbf{21.5\%}&\textbf{4.6}\\
      \bottomrule
    \end{tabular}
    \label{table:ablation}
    \vspace{-2.6em}
  \end{center}
\end{table}

\textbf{Evaluation of the model  with different designs.}
In this section, we perform extensive ablation studies on the SURREAL dataset to evaluate the effectiveness of each design.
Table~\ref{table:ablation} demonstrates the performance of the model with different designs.
Specifically, the first line is the results of DPC~\cite{lang2021dpc}, which is our baseline model. 
The second row indicates that the self-ensembling framework with a stochastic transform achieves a better performance than the original model.
By using $\mathcal{L}_{css}$ to constrain the consistency of source features before and after augmentation, the correspondence accuracy can be improved by 0.4\%, as shown in the third row.
As shown in the fourth row, adding the Orientation Evaluation Module without the Feature Interaction Module and the Domain Adaptation Module, the performance has a slight improvement.
In the fifth row, by introducing the Feature Interaction Module into the Orientation Evaluation Module, the correspondence accuracy can be improved by 0.9\%.
Finally, after utilizing the Domain Adaptation Module, the correspondence accuracy is improved by 1.1\%.
\begin{table}[!t]
  \begin{center}
      \footnotesize
      \setlength\tabcolsep{6pt}
      \caption{\textbf{Effect of the Self-Ensembling Framework on SURREAL.} Here, we modify the self-ensembling framework to show the effect of each designed component. $\mathcal{N}$ means Gaussian noise, and $\mathcal{R}$ means random rotation along the vertical z-axis.}
      \label{table:Self-Ensemble}
      \vspace{-0.8em}
      \begin{tabular}{c|c|c|c|c}
          \toprule
          SURREAL &w/o $\mathcal{L}_{ccs}$& w/o $\mathcal{L}_{css}$& w/o $\mathcal{N}$&w/o $\mathcal{R}$\\
          \midrule
          acc $\uparrow$ & 19.9\% & 20.3\% & 20.6\% &18.8\% \\
          err $\downarrow$ & 5.3 & 5.1 & 4.9 &5.7 \\
          \bottomrule
      \end{tabular}
      \vspace{-2em}
  \end{center}
\end{table}

\begin{figure}[!t]
  \vspace{-0.5em}
  \begin{center}
      \includegraphics[width=0.36\textwidth]{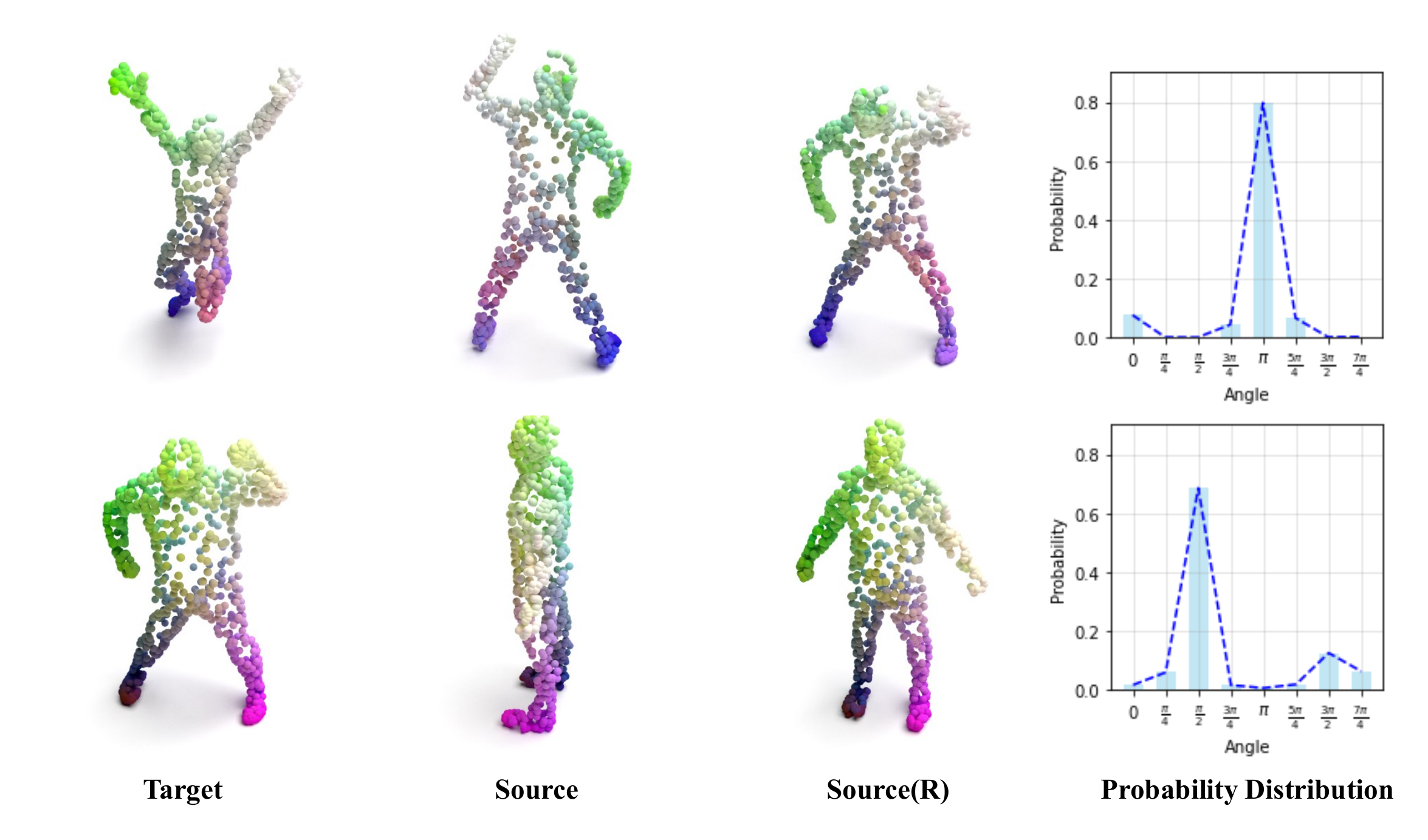}
      \vspace{-1.4em}
      \caption{\textbf{Effect of the Orientation Estimation Module.} We visualize of the point clouds before and after orientation rectification on SHREC'19 test set. We denote the raw source point cloud as Source and the rectified one as Source(R). Besides, we provide the probability distributions of the relative rotation angle prediction.}
      \vspace{-2.0em}
      \label{orientation_est}
  \end{center}
\end{figure}

\textbf{Effect of the Self-Ensembling Framework.}
As shown in Table~\ref{table:Self-Ensemble}, we modify the self-ensembling framework to show the effect of each designed component.
Removing either of the consistency losses leads to a drop in performance, which indicates that constraining the student network with soft labels facilitates the consistency of the point cloud representations. 
When Gaussian noise and rotation augmentation are removed, the model performances show different degrees of degradation. 
%
%
The above experiments illustrate our self-ensemble method can obtain a more robust feature representation of the point cloud through data augmentation and consistency losses.
%


\textbf{Effect of the Orientation Estimation Module.}
To verify the effect of the proposed Orientation Estimation Module, we visualize the point clouds after orientation rectification and the probability distributions of the relative rotation angle predictions.
As shown in Figure~\ref{orientation_est}, our method can accurately estimate the relative rotation angle and align the point cloud orientation of the source with that of the target.
More results refer to the supplementary
materials.

\subsection{Robustness Analysis}
\label{Robustness_Analysis}

\begin{table}[!t]
  \begin{center}
    \footnotesize
    \setlength\tabcolsep{2pt}
    \centering
    \caption{\textbf{Robustness Analysis.}
    To verify the robustness of our method, we test our method and baselines using different augmented test sets. $\mathcal{N}$ means Gaussian noise with standard deviation $\sigma$ of 0.1 and $\mathcal{R}$ means random rotation along the vertical z-axis.}
    \vspace{-0.9em}
    \begin{tabular}{cc|cc|cc|cc|cc}
      \toprule
      \multirow{2}*{$\mathcal{N}$} &  \multirow{2}*{$\mathcal{R}$} & \multicolumn{2}{c|}{SURREAL(B)} & \multicolumn{2}{c|}{SURREAL} & \multicolumn{2}{c|}{SMAL(B)} & \multicolumn{2}{c}{SMAL} \\
      \cline{3-10}
      & &acc $\uparrow$ & err $\downarrow$ & acc $\uparrow$ & err $\downarrow$ &acc $\uparrow$ & err $\downarrow$ & acc $\uparrow$ & err $\downarrow$\\
      \midrule
      \ding{55}&\ding{55}&17.47\%&6.30&21.55\%&4.65&33.79\%&5.78&36.39\%&3.88\\
      \ding{51}&\ding{55}&14.38\%&8.74&21.55\%&4.66&30.21\%&6.06&36.15\%&3.92\\
      \ding{55}&\ding{51}&8.99\%&9.33&17.59\%&5.81&9.98\%&12.24&28.16\%&5.63\\
      \ding{51}&\ding{51}&7.80\%&11.28&17.58\%&5.79&9.54\%&12.80&27.60\%&5.89\\
      \bottomrule
    \end{tabular}
    \label{table:robustness}
    \vspace{-2.5em}
  \end{center}
\end{table}

\begin{figure}[!t]
  \begin{center}
    \vspace{-1.8em}
      \includegraphics[width=0.36\textwidth]{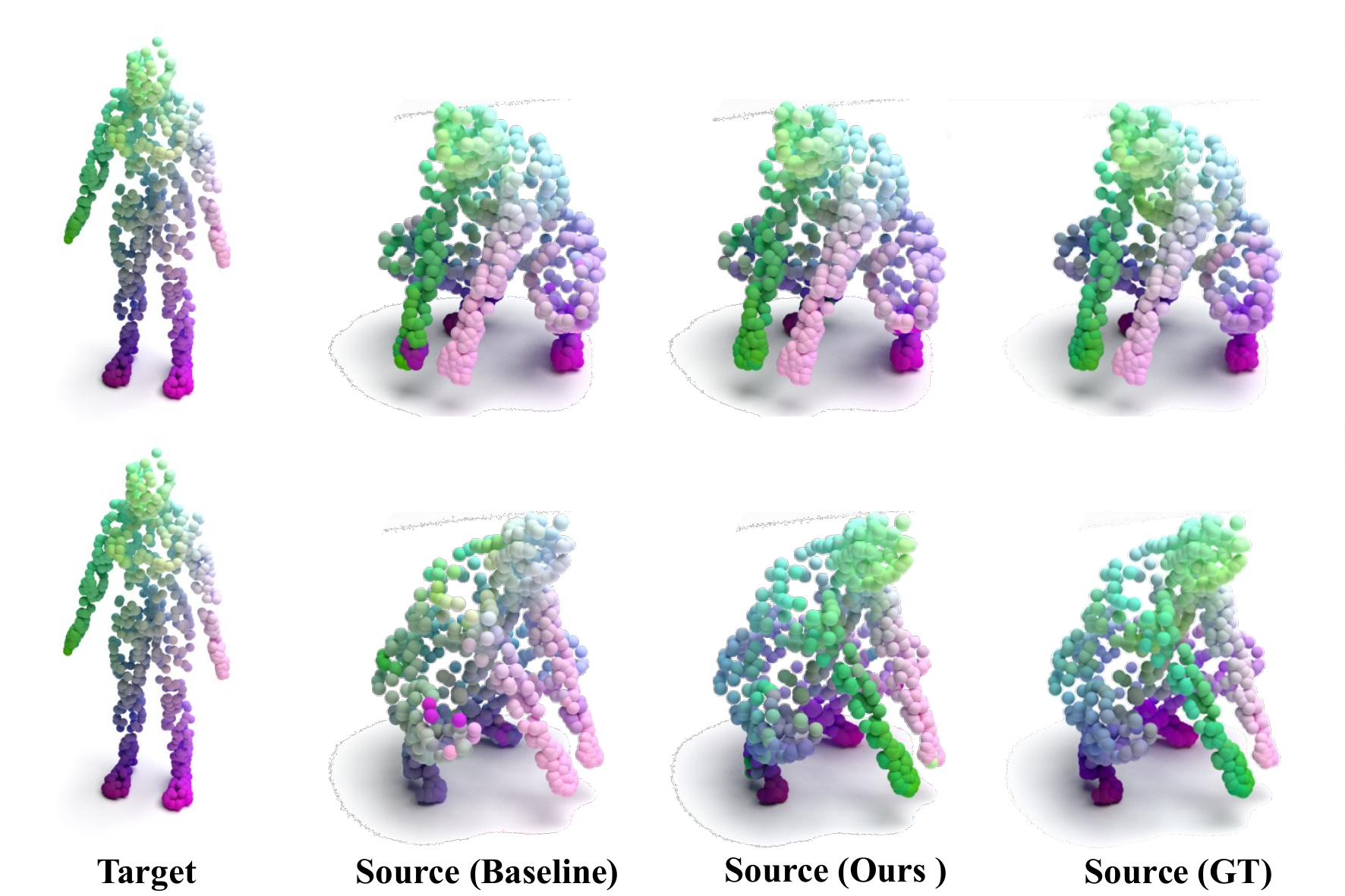}
      \vspace{-0.8em}
      \caption{\textbf{Visualization of the correspondence results with rotation augmentation. }
      With rotation augmentation on point cloud pairs, the baseline shows regrettable performances, while our method still retains desirable performances.
      }
      \vspace{-1.8em}
      \label{rotation_samples}
  \end{center}
\end{figure}

\begin{figure}[!t]
  \begin{center}
      \includegraphics[width=0.38\textwidth]{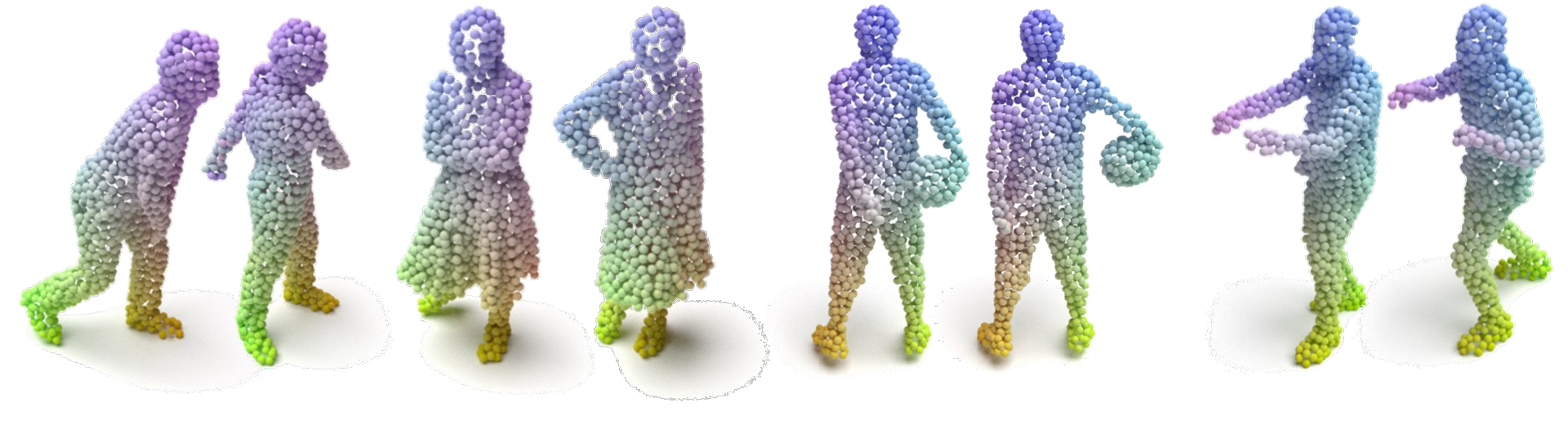}
      \vspace{-1.8em}
      \caption{\textbf{Visualization of point cloud pairs with deformations on real scanned Owlii dataset.}
      In each pair, the left one is the source and the right one is the target.
      }
      \vspace{-2.5em}
      \label{real_dataset}
  \end{center}
\end{figure}

To verify the robustness of our method, we use different augmentations on the test set. As shown in Table~\ref{table:robustness}, we use Gaussian noise and random rotation for data augmentation of the test sets. 
Compared to our method, the baseline shows a significant performance decrease for test sets with Gaussian noise.
For test sets with random rotation, the baseline performance is greatly degraded, while our method retains an acceptable performance.
Figure~\ref{rotation_samples} shows that our SE-ORNet can handle the orientation inconsistency issue of source and target well.
To further verify the robustness and generalization of our method, we conduct experiments on the real scanned Owlii dataset~\cite{xu2017owlii} and present the visualization in Figure~\ref{real_dataset}.
The results show that our SE-ORNet trained on the synthetic SURREAL dataset still produces impressive performance on the real scanned dataset, demonstrating strong generalization and robustness. 



%% file: clu.tex
In this paper, we propose a self-ensembling orientation-aware network for unsupervised point cloud shape correspondence.
To the best of our knowledge, SE-ORNet is the first self-ensembling network in this area.
To solve the mismatching issue of symmetrical parts in point clouds with different body orientations, we design a plug-and-play lightweight orientation estimation module to align the orientations of two point clouds.
Extensive experiments conducted on four shape correspondence benchmarks demonstrate the superior performance of   SE-ORNet.